\documentclass[
]{ceurart}

\usepackage{natbib}
\usepackage{algorithm}
\usepackage{algorithmic}

\usepackage{amsmath,amssymb,amsfonts}
\usepackage{booktabs}
\usepackage{subfigure}
\usepackage{hyperref}

\begin{document}

\copyrightyear{2022}
\copyrightclause{Copyright for this paper by its authors.
  Use permitted under Creative Commons License Attribution 4.0
  International (CC BY 4.0).}

\conference{De-Factify: Workshop on Multimodal Fact Checking and Hate Speech Detection, co-located with AAAI 2022. 2022 Vancouver, Canada}

\title{Team Yao at Factify 2022: Utilizing Pre-trained Models and Co-attention Networks for Multi-Modal Fact Verification}

\author[1]{Wei-Yao Wang}[%
email=sf1638.cs05@nctu.edu.tw,
]
\address[1]{Department of Computer Science, National Yang Ming Chiao Tung University, Hsinchu, Taiwan}

\author[1]{Wen-Chih Peng}[%
email=wcpeng@nctu.edu.tw,
]

\begin{abstract}
In recent years, social media has enabled users to get exposed to a myriad of misinformation and disinformation; thus, misinformation has attracted a great deal of attention in research fields and as a social issue.
To address the problem, we propose a framework, Pre-CoFact, composed of two pre-trained models for extracting features from text and images, and multiple co-attention networks for fusing the same modality but different sources and different modalities.
Besides, we adopt the ensemble method by using different pre-trained models in Pre-CoFact to achieve better performance.
We further illustrate the effectiveness from the ablation study and examine different pre-trained models for comparison.
Our team, Yao, won the fifth prize (F1-score: 74.585\%) in the Factify challenge hosted by De-Factify @ AAAI 2022, which demonstrates that our model achieved competitive performance without using auxiliary tasks or extra information.
The source code of our work is publicly available\footnote{https://github.com/wywyWang/Multi-Modal-Fact-Verification-2021}.
\end{abstract}


\maketitle

\section{Introduction}

Fake news has become easier to spread due to the growing number of users of social media.
For example, about 59\% of social media consumers expect that news spread via social media may be inaccurate \cite{Shearer21}.
To influence social thoughts, there are many fake news stories that mislead readers about the news content by replacing some true content with false details.
Besides, fake news with textual and visual content can better attract readers and it is hard to judge than only using textual content.
Therefore, it is essential to detect multi-modal fake news to eliminate its negative impacts.

Fake checkers aim to check the worthiness, evidence or verified claim retrieval \cite{DBLP:conf/ijcai/NakovCHAEBPSM21}.
Recent works have presented a number of approaches for tackling fake news detection automatically.
In uni-modal detection, \citet{DBLP:conf/wsdm/ShuWL19} exploited a tri-relationship (publishers, news pieces, and users) to model the relations and interactions for detecting news disinformation.
\citet{DBLP:conf/aaai/Przybyla20} utilized the style the news articles are written in to estimate their credibility.
In multi-modal detection, \citet{DBLP:conf/mm/JinCGZL17} proposed an att-RNN that combines a recurrent neural network with an attention mechanism to fuse textual content and visual images.
MCAN is proposed by extracting spatial-domain features and textual features by pre-trained models \cite{DBLP:conf/acl/WuZZWX21}.
Further, to address the fact that fake images are often re-compressed images or tampered images, which shows periodicity in the frequency domain, they used discrete cosine transform as in \cite{DBLP:conf/icdm/QiCYGL19}, then designed a CNN-based network for capturing frequency-domain features from images.

\begin{figure*}
  \centering
  \includegraphics[width=\linewidth]{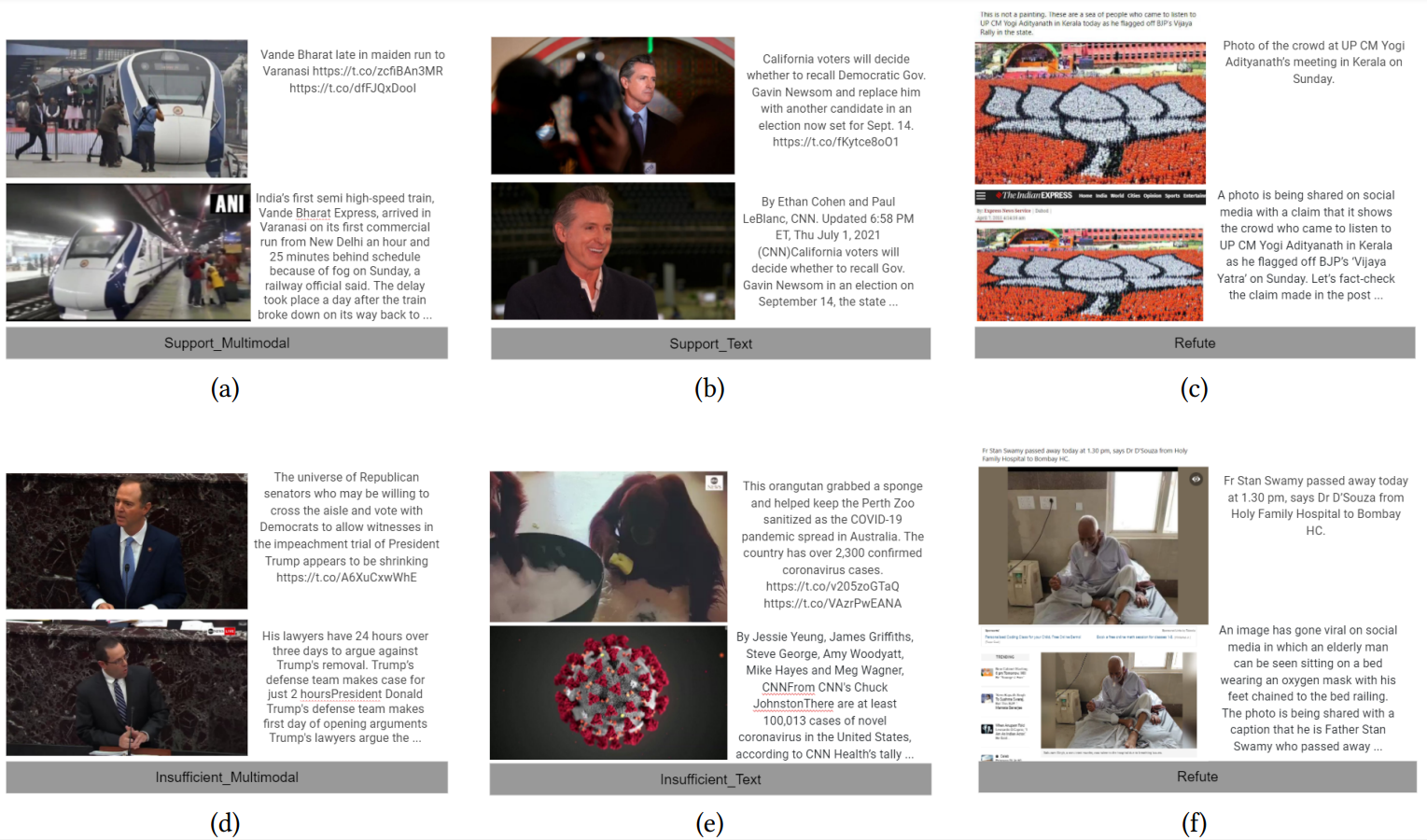}
  \caption{A screenshot from \cite{defactify21}, which illustrates sample examples of five categories from the Factify dataset.}
  \label{fig:introduction-example}
\end{figure*}

A real-world problem, identifying if the claim entails the document, is the challenge called Factify \cite{defactify21} hosted by De-Factify\footnote{https://aiisc.ai/defactify/factify.html}.
Figure \ref{fig:introduction-example} shows some examples for all five categories.
The goal is to design a method to classify the given text and images into one of the five categories: Support\_Multimodal, Support\_Text, Insufficient\_Multimodal, Insufficient\_Text, and Refute.
To tackle the problem, in this paper, we propose Pre-CoFact with pre-trained models and co-attention networks to perform the shared task, which first extracts features from both text and images, then fuses this information through the co-attention module.
Specifically, two powerful Transformer-based pre-trained models, DeBERTa \cite{DBLP:conf/iclr/HeLGC21} and DeiT \cite{DBLP:conf/icml/TouvronCDMSJ21}, are adopted for extracting features from both claims and documents' text and images, respectively.
Afterwards, several co-attention modules are designed for fusing the contexts of the text and images.
Finally, these embeddings are aggregated as corresponding embeddings to classify the category of the news.

The main results of this paper can be summarized as follows:
\begin{itemize}
    \item Using text and images directly can achieve expressive results without any auxiliary tasks, preprocessing methods, or extra information (\textit{e.g.}, optical character recognition (OCR) from images).
    \item Adopting pre-trained models helps improve the performance of the shared task, and using co-attention networks can learn the correlation from the same modality (text or images from claims and documents) and the dependencies between different modalities (text and images).
    \item Our ensemble model outperforms the machine learning models \cite{defactify21} at least 48\% and 40\% in terms of validation score and testing score.
    Besides, extensive experiments were further conducted to examine the capability of the proposed model.
\end{itemize}
\section{Dataset}
Factify is a dataset for multi-modal fact verification, which contains images, textual claims, reference textual documents and images.
Each sample includes claim\_image, claim, claim\_ocr, document\_image, document, document\_ocr, and category.
The detail of each field is described as follows:
\begin{itemize}
    \item claim\_image: the image of the given claim.
    \item claim: the text of the given claim.
    \item claim\_ocr: the text from the claim\_image detected by the host.
    \item document\_image: the image of the given reference.
    \item document: the text of the given reference.
    \item document\_ocr: the text from the document\_image detected by the host.
    \item category: the category of the data sample from a list of five classes.
\end{itemize}
The category is composed of 1) Support\_Multimodal: both the claim text and image are similar to that of the document, 2) Support\_Text: the claim text is similar or entailed, but images of the document and claim are not similar, 3) Insufficient\_Multimodal: the claim text is neither supported nor refuted by the document but images are similar to the document, 4) Insufficient\_Text: both text and images of the claim are neither supported nor refuted by the document, although it is possible that the text claim has common words with the document text, and 5) Refute: the images and/or text from the claim and document are completely contradictory.

The training set contains 35,000 samples, which has 5,000 samples of each class, and the validation set contains 7,500 samples, which has 1,500 samples of each class.
The test set, which is used to evaluate the private score, also contains 7,500 samples.
For more details, we refer readers to \cite{defactify21}.
\section{Related Works}

\subsection{Fake News Detection}
There have been a series of studies combating fake news detection to mitigate a societal crisis \cite{DBLP:conf/kdd/NakovM21}.
\citet{DBLP:conf/emnlp/VoL20} proposed a novel neural ranking model which jointly utilizes textual and visual matching signals.
This is the first work using multi-modal data in social media posts to search for verified information, which can increase users' awareness of fact-checked information when they are exposed to fake news.
\citet{DBLP:conf/naacl/LeeBMF21} adopted a perplexity-based approach in the few-shot learning, which assumes that the given claim may be fake if the corresponding perplexity score from evidence-conditioned language models is high.
BertGCN \cite{DBLP:journals/corr/abs-2105-05727} is proposed by integrating the advantages of large-scale pre-trained models and graph neural networks for fake news detection, which is able to learn the representations from the massive amount of pre-trained data and the label influence through the propagation.
MCAN \cite{DBLP:conf/acl/WuZZWX21} adopts a large-scale pre-trained NLP model and a pre-trained computer vision (CV) model for extracting features from text and images, respectively.
Besides, MCAN also extracts frequency domain features from images, and then uses multiple co-attention layers to fuse this information.

These approaches demonstrate the effectiveness of using pre-trained models for fake news detection, which motivated us to use pre-trained models as well.
Besides, MCAN inspires us to fuse the contexts of different modalities or the same modality (\textit{e.g.}, text from claims and documents).

\subsection{Large-Scale Pre-trained Models}
Transformer \cite{DBLP:conf/nips/VaswaniSPUJGKP17} has been used for machine translation and has inspired many competitive approaches in natural language processing (NLP) tasks.
Transformer-based pre-trained language models (PLMs) have significantly improved the performance of various NLP tasks due to the ability to understand contextualized information from the pre-trained dataset.
Since BERT \citep{DBLP:conf/naacl/DevlinCLT19} was presented, we have seen the rise of a set of large-scale PLMs such as GPT-3 \citep{DBLP:conf/nips/BrownMRSKDNSSAA20}, RoBERTa \citep{DBLP:journals/corr/abs-1907-11692}, XLNet \citep{DBLP:conf/nips/YangDYCSL19}, ELECTRA \citep{DBLP:conf/iclr/ClarkLLM20}, and DeBERTa \citep{DBLP:conf/iclr/HeLGC21}.
These PLMs have been fine-tuned using task-specific labels and have created a new state of the art in many downstream tasks.

Recently, vision Transformer (ViT) \cite{DBLP:conf/iclr/DosovitskiyB0WZ21} is a Transformer encoder architecture directly applied to image classification with patching raw images as input to NLP, which achieves competitive results compared to state-of-the-art convolutional networks by pre-training a large private image dataset JFT-300M \cite{DBLP:conf/iccv/SunSSG17}.
ViT demonstrates that convolution-free networks can still learn the relation in the images.
To reduce the pre-trained dataset size and training efficiency, several follow-up studies have been conducted.
DINO was proposed by \cite{DBLP:journals/corr/abs-2104-14294} to improve the standard ViT model through self-supervised learning.
\cite{DBLP:conf/icml/TouvronCDMSJ21} proposed DeiT, which used a novel distillation procedure based on a distillation token to ensure the student learns from the teacher through attention.

These pre-trained models demonstrate the generalization of various domains.
Further, using pre-trained models benefits capturing rich information of downstream tasks, which can also reduce the burden of training from scratch.
These advantages motivated us to adopt state-of-the-art pre-trained models for transforming images and text into contextual embeddings.
Besides, we focused on using Transformer-based pre-trained models for feature extraction.
\section{Method}
\subsection{Problem Formulation}
Let $C=\{CT_i, CI_i, DT_i, DI_i\}^{|C|}_{i=1}$ denote the corpus of the dataset, where the $i$-th sample is composed of the claim text $CT_i=w_1^{CT_i}w_2^{CT_i}\cdots$, the claim image $CI_i$, the document text $DT_i=w_1^{DT_i}w_2^{DT_i}\cdots$, and the document image $DI_i$.
The $i$-th target $y_i \in \{Support\_Multimodal, Support\_Text,\allowbreak Insufficient\_Multimodal, Insufficient\_Text, Refute\}$.
The goal is to find out support, insufficient-evidence and refute between given claims and documents.

\begin{figure*}
  \centering
  \includegraphics[width=\linewidth]{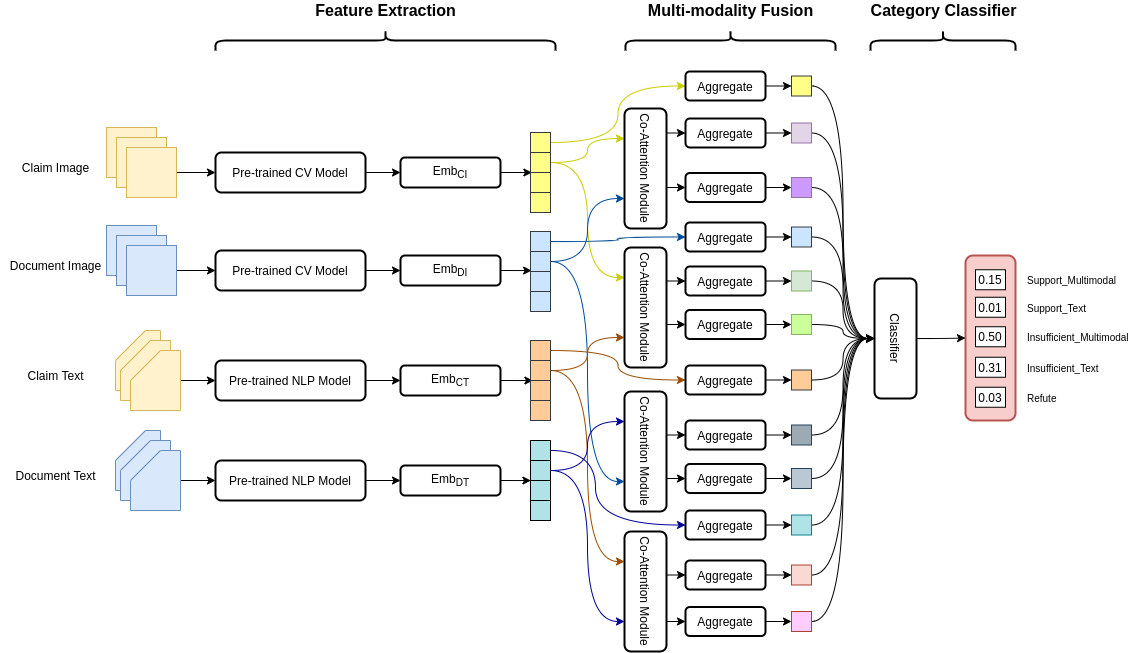}
  \caption{Illustration of the Pre-CoFact framework.
  Each square can be seen as a token with a $d$ dimension vector.
  The feature extraction part aims to transform text and images into corresponding embeddings.
  The multi-modality fusion part fuses this information from the same modality (images/text from the claim and document) and different modalities (images and text from the claim/document) to obtain contexts.
  Finally, the category classifier predicts the possible classes based on the embeddings from feature extraction and the embeddings from multi-modality fusion.}
  \label{fig:framework-overview}
\end{figure*}

\subsection{Pre-CoFact Overview}
Figure \ref{fig:framework-overview} illustrates the overview of the proposed Pre-CoFact framework.
The input contains the claim image, the claim text, the document image, and the document text.
The feature extraction part adopts DeiT \cite{DBLP:conf/icml/TouvronCDMSJ21} as the pre-trained CV model and DeBERTa \cite{DBLP:conf/iclr/HeLGC21} as the pre-trained NLP model, and feeds the outputs of pre-trained models to the image embedding layer and text embedding layer for transforming images and texts into corresponding embeddings.
The multi-modality fusion part fuses this information from the same modality (images/text from the claim and document) and different modalities (images and text from the claim/document) based on multiple co-attention layers.
Finally, the category classifier predicts the possible classes based on the embeddings from feature extraction and the embeddings from multi-modality fusion.

\subsection{Feature Extraction}
The enrichment of pre-trained models enables us to have rich information without training from scratch.
Moreover, Transformer-based pre-trained models demonstrate the success on both NLP and CV tasks.
However, it is essential to fine-tune for fitting in our task.
To this end, we first use DeBERTa as our pre-trained NLP model and DeiT as our pre-trained CV model, and then we use the embedding layer for transforming pre-trained embeddings to embeddings in our task.
Specifically, the $i$-th output of the embedding layer is calculated as follows:
\begin{equation}
    E_{CI_i} = Emb_{CI}~(DeiT(CI_i)), E_{DI_i} = Emb_{DI}~(DeiT(DI_i)),
\end{equation}
\begin{equation}
    E_{CT_i} = Emb_{CT}~(DeBERTa(CT_i)), E_{DT_i} = Emb_{DT}~(DeBERTa(DT_i)),
\end{equation}
where the output dimensions of DeiT and DeBERTa are 768, the $Emb$ is composed of a MLP and an activation function, and $E_{CI_i}, E_{CT_i}, E_{DI_i}, E_{DT_i}$ are $d$ dimension vectors.
It is noted that the activation functions in $Emb$ we used are ReLU and Mish \cite{DBLP:journals/corr/abs-1908-08681} for testing the results.

\subsection{Multi-Modality Fusion}
After generating embeddings of text and images, we adopt multiple co-attention layers as in \cite{DBLP:conf/acl/WuZZWX21} to fuse the embeddings.
To check the relation between claim and document, we use the co-attention layer to separately fuse 1) images of claims and documents and 2) text of claims and documents.
Besides, the relation between text and images from the claims or document can be viewed as checking whether they are relative or not.
Therefore, we also adopt the co-attention layer for fusing 3) images and text of claims and 4) images and text of documents.

Specifically, each co-attention layer takes two inputs $E_A$ and $E_B$ and produces two outputs $H_A, H_B$.
Here we use a single head to derive as the following equations:
\begin{equation}
    Q_A = E_AW^{Q_A}, K_A = E_AW^{K_A}, V_A = E_AW^{V_A}, Q_B = E_BW^{Q_B}, K_B = E_BW^{K_B}, V_B = E_BW^{V_B},
\end{equation}
\begin{equation}
    \tilde{H_A} = Norm(E_A + softmax(\frac{Q_A K_B^T}{\sqrt{d}})V_B), \tilde{H_B} = Norm(E_B + softmax(\frac{Q_B K_A^T}{\sqrt{d}})V_A),
\end{equation}
\begin{equation}
    H_A = Norm(\tilde{H_A} + FFN(\tilde{H_A})), H_B = Norm(\tilde{H_B} + FFN(\tilde{H_B})),
\end{equation}
where $W^{Q_A}, W^{K_A}, W^{V_A}, W^{Q_B}, W^{K_B}, W^{V_B} \in \mathbb{R}^{d \times d}$, and $Norm$ and $FFN$ is the same normalization method and feed forward network as in \cite{DBLP:conf/nips/VaswaniSPUJGKP17}.
Co-attention block has been widely used in VQA tasks \cite{DBLP:conf/cvpr/GaoJYLHWL19}, as it can capture dependencies of different inputs.
Therefore, we use the co-attention layer for fusing:
\begin{equation}
    H_{CIDI_i}, H_{DICI_i} = CoAtt(E_{CI_i}, E_{DI_i}), H_{CTDT_i}, H_{DTCT_i} = CoAtt(E_{CT_i}, E_{DT_i}),
\label{same-modality}
\end{equation}
\begin{equation}
    H_{CIDT_i}, H_{DTCI_i} = CoAtt(E_{CI_i}, E_{DT_i}), H_{CTDI_i}, H_{DICT_i} = CoAtt(E_{CT_i}, E_{DI_i}),
\end{equation}
where $CoAtt$ denotes the co-attention layer.

After applying the co-attention mechanism, the aggregation function is adopted to aggregate fused tokens into a representative token.
That is, given a fused embedding with $\mathbb{R}^{N \times d}$, where $N$ is the sequence length, we use mean aggregation to output $\mathbb{R}^{1 \times d}$.
Besides, we also feed $E_{CI_i}, E_{CT_i}, E_{DI_i}, E_{DT_i}$ into the aggregation function for classification.

\subsection{Category Classifier}
To predict the label of the given claims and documents, we first concatenate 8 aggregated outputs $H_{CIDI_i}$, $H_{DICI_i}$, $H_{CTDT_i}$, $H_{DTCT_i}$, $H_{CIDT_i}$, $H_{DTCI_i}$, $H_{CTDI_i}$, $H_{DICT_i}$ from the co-attention layers and original 4 aggregated embeddings $E_{CI_i}, E_{CT_i}, E_{DI_i}, E_{DT_i}$ to obtain the input of the classifier $Z_i$.
It is worth noting that the outputs of embeddings are also used since the original information can provide some clues for classifying the news.
Afterwards, the $i$-th output of the classifier is the probability as follows:
\begin{equation}
    Z_i^{M_1} = \sigma(Z_i W^{Z}), Z_i^{M_2} = \sigma(Z_i^{M_1} W^{M_1}),
\end{equation}
\begin{equation}
    \hat{y_i} = softmax(Z_i^{M_2} W^{M_2}),
\end{equation}
where $W^{Z} \in \mathbb{R}^{12d \times d}$, $W^{M_1} \in \mathbb{R}^{d \times d_{M_1}}$, and $W^{M_1} \in \mathbb{R}^{d_{M_1} \times 5}$.
Note that $\sigma$ is the same as in $Emb$, which uses both ReLU and Mish for testing the results.

We trained our model by minimizing cross-entropy loss $\mathbb{L}$ to learn the prediction of the categories:
\begin{equation}
    \mathbb{L} = -\sum_{i=1}^{|C|} y_i log(\hat{y}_i).
\end{equation}

\subsection{Ensemble Method}
Each classifier may have its strengths and weakness, and ensemble methods have been widely used to enhance the performance.
Therefore, we follow \cite{DBLP:journals/corr/abs-2007-02259} to use the power weighted sum to enhance the performance of the model.
The formula is derived as follows:
\begin{equation}
    p = p_1^N \times w_1 + p_2^N \times w_2 + \cdots + p_k^N \times w_k,
\end{equation}
where $p_i,\cdots,p_k$ are the predicted probability from the corresponding model, $w_1,\cdots,w_k$ are weights with respect to the corresponding model, $k$ is the number of trained models, and $N$ is the weight of power.
It is noted that these parameters are tuned by hand.
\section{Results and Analysis}

\subsection{Experimental Setup}
\subsubsection{Implementation Details}
The dimension $d$ was set to 512, the inner dimension of the feed-forward layer was 1024, and the number of heads was set to 4.
The dropout rate was 0.1, and the max sequence length was 512.
The batch size was 32, the learning rates were set to 3e-5 and 2e-5, the training epochs were set to 30, and the seeds were tested with 41 and 42.
The power $N$ was set to 0.5, and the weights were set to 0.6, 0.2, 0.1, 0.2, 0.3, which were manually tuned by validation score.
The pre-trained DeBERTa was deberta-base\footnote{https://huggingface.co/microsoft/deberta-base}, and the DeiT was deit-base-patch16-224\footnote{https://huggingface.co/facebook/deit-base-patch16-224}.
The parameters of the two pre-trained models were frozen.
All images were transformed by resizing to 256, center cropping to 224, and normalizing.
We preprocessed only for transforming images, and then we stored the text and processed images in corresponding pickle files for training and evaluating.
All the training and evaluation phases were conducted on a machine with Intel Xeon 4110 CPU @ 2.10GHz, Nvidia GeForce RTX 2080 Ti, and 252GB RAM.
The source code is available at https://github.com/wywyWang/Multi-Modal-Fact-Verification-2021.

\subsubsection{Evaluation Metric}
To evaluate the performance of the task, the weighted average F1 score was used across the 5 categories.

\subsection{Quantitative Results}
\subsubsection{Ablation Study}
We first conducted an ablation study to ensure the effective design of our proposed Pre-CoFact.
As shown in Table \ref{tab:experiment-ablation}, it is evident that without co-attention networks (w/o CoAtt), the performance is degraded.
Further, applying co-attention only on the same modality (w/o CoAtt(text, image)) is insufficient, which demonstrates the need for modeling dependencies between different modalities.
It is noted that our ensemble method slightly improves the performance compared to Pre-CoFact.
Our ensemble method includes Pre-CoFact, Pre-CoFact with replacing DeBERTa with XLM-RoBERTa, Pre-CoFact with replacing DeBERTa with RoBERTa, Pre-CoFact with replacing DeBERTa with RoBERTa and replacing ReLU with Mish, and Pre-CoFact with replacing ReLU with Mish.

We also use different pre-trained models to examine the module influence as shown in Table \ref{tab:experiment-variant}.
It can be seen that DeiT is more suitable than DINO for this task.
Besides, XLM-RoBERTa also degrades the performance, while RoBERTa is slightly worse than Pre-CoFact with DeBERTa.

\begin{table}
  \centering
  \begin{tabular}{c||cc||cc}
    \toprule
    Model & w/o CoAtt & w/o CoAtt(text, image) & Pre-CoFact (Ours) & Ensemble (Ours) \\
    \midrule
    Weighted F1 (\%) & 75.68 (-4.34) & 76.43 (-3.59) & 78.46 & \textbf{80.02} (+1.56) \\
    \bottomrule
\end{tabular}
  \caption{Ablation study of our model in terms of validation score. w/o CoAtt denotes using four embeddings for classification and w/o CoAtt(text, image) denotes using only the same modality (Equ. \ref{same-modality}).}
  \label{tab:experiment-ablation}
\end{table}

\begin{table}
  \centering
  \begin{tabular}{c||ccc||c}
    \toprule
    Model & DINO \cite{DBLP:journals/corr/abs-2104-14294} & XLM-RoBERTa \cite{DBLP:conf/acl/ConneauKGCWGGOZ20} & RoBERTa \cite{DBLP:journals/corr/abs-1907-11692} & Pre-CoFact (Ours) \\
    \midrule
    Weighted F1 (\%)             & 73.94 (-4.52) & 74.11 (-4.35) & 77.53 (-0.93) & 78.46 \\
    \bottomrule
\end{tabular}
  \caption{Variant pre-trained models in terms of validation score. Pre-CoFact uses DeiT and DeBERTa as pre-trained models. DINO is replaced DeiT by DINO \cite{DBLP:journals/corr/abs-2104-14294}. XLM-RoBERTa and RoBERTa are replaced DeBERTa by XLM-RoBERTa \cite{DBLP:conf/acl/ConneauKGCWGGOZ20} and RoBERTa \cite{DBLP:journals/corr/abs-1907-11692}, respectively.}
  \label{tab:experiment-variant}
\end{table}

\begin{table}
  \centering
  \addtolength{\tabcolsep}{-2pt}
  \begin{tabular}{cc||ccccc||c}
    \toprule
    & & Support & Support & Insufficient & Insufficient & & \\
    Rank & Team & \_ Text (\%) & \_ Multimodal (\%) & \_Text (\%) & \_Multimodal (\%) & Refute (\%) & Final (\%) \\
    \midrule
    5 & \textbf{Yao}             & 68.881 & \textbf{81.610} & \textbf{84.836} & \textbf{88.309} & \textbf{100.00} & \textbf{74.585} \\
    - & Baseline             & 82.675 & 75.466 & 74.424 & 69.678 & 42.354 & 53.098 \\
    \bottomrule
\end{tabular}
  \caption{Performance of our model in terms of testing score. Our method achieved fifth prize with only about a 2.2\% gap, while we outperformed the baseline by 40.5\%.}
  \label{tab:experiment-test}
\end{table}

\subsubsection{Testing Performance}
Table \ref{tab:experiment-test} shows the performance of the testing set.
Our approach achieved 74.585\% of the F1-score, winning the fifth prize in detecting fake news.
This result outperformed the baseline by 40.5\%, while it still has only a 2.2\% gap compared to the first prize.
Despite the result, our approach still demonstrates that using only text and images can achieve competitive performance.

\subsubsection{Confusion Matrix}
\begin{figure*}
  \centering
  \includegraphics[width=\linewidth]{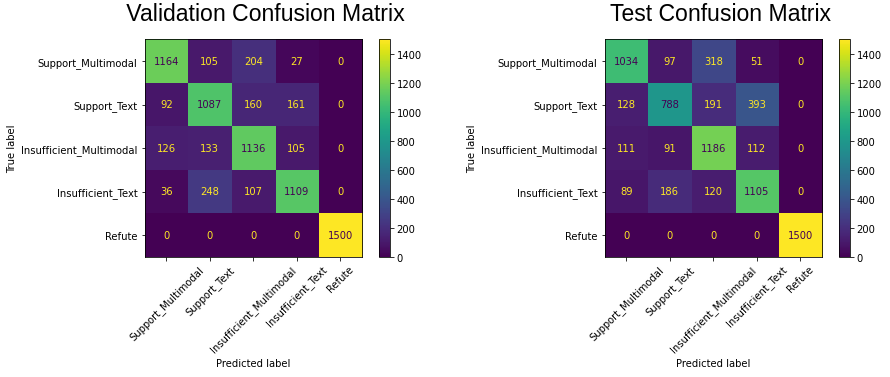}
  \caption{Confusion matrix of the validation set and testing set.}
  \label{fig:experiment-confusion}
\end{figure*}

Figure \ref{fig:experiment-confusion} shows the confusion matrix of the validation set and testing set.
It can be observed that our model can precisely classify refute on both the validation set and testing set, while our model misjudged whether the text is entailed when the image is not entailed.

\section{Conclusion}
In this paper, we proposed Pre-CoFact utilizing pre-trained models and multiple co-attention networks to alleviate the effect of fake news for the Factify task.
To achieve better performance, we adopted an ensemble method by weighting several models.
The ablation study demonstrates the effectiveness of our proposed approach.
From the testing score, our method illustrates that using only text and images without extra information can also achieve competitive performance.

\bibliography{reference}

\end{document}